\documentclass[runningheads]{llncs}
\usepackage[T1]{fontenc}
\usepackage{graphicx}
\usepackage{xcolor}
\usepackage{amsmath} 
\begin{document}
\title{Utilizing Large Language Models for Machine Learning Explainability}

\author{Alexandros Vassiliades\inst{1}\orcidID{0000-0003-4569-503X} \and
     Nikolaos Polatidis\inst{2}\orcidID{0000-0003-4249-4953} \and Stamatios Samaras\inst{1}\orcidID{0009-0000-0010-2064} \and Sotiris Diplaris\inst{1}\orcidID{0000-0002-9969-6436} \and Ignacio Cabrera Martin\inst{2}\orcidID{0009-0006-4729-6795} \and Yannis Manolopoulos\inst{3}\orcidID{0000-0003-4026-4329}  \and Stefanos Vrochidis\inst{1}\orcidID{0000-0002-2505-9178} \and Ioannis Kompatsiaris\inst{1}\orcidID{0000-0001-6447-9020}}

 \authorrunning{Alexandros Vassiliades et al.}

 \institute{Centre for Research \& Technology, Hellas \and
 School of Architecture Technology and Engineering, 
 University of Brighton, BN2 4GJ, United Kingdom \and 
 University of Nicosia, 2417 Cyprus \\
 \email{valexande@iti.gr}}
\maketitle 
\begin{abstract}
This study explores the explainability capabilities of large language models (LLMs), when employed to autonomously generate machine learning (ML) solutions. We examine two classification tasks: 
(i) a binary classification problem focused on predicting driver alertness states, and 
(ii) a multilabel classification problem based on the yeast dataset. 
Three state-of-the-art LLMs (i.e. OpenAI GPT, Anthropic Clau\-de, and DeepSeek) are prompted to design training pipelines for four common classifiers: Random Forest, XGBoost, Multilayer Perceptron, and Long Short-Term Memory networks. The generated models are evaluated in terms of predictive performance (recall, precision, and F1-score) and explainability using SHAP (SHapley Additive exPlanations). Specifically, we measure Average SHAP Fidelity (Mean Squared Error between SHAP approximations and model outputs) and Average SHAP Sparsity (number of features deemed influential). The results reveal that LLMs are capable of producing effective and interpretable models, achieving high fidelity and consistent sparsity, highlighting their potential as automated tools for interpretable ML pipeline generation. The results show that LLMs can produce effective, interpretable pipelines with high fidelity and consistent sparsity, closely matching manually engineered baselines.

\end{abstract}

\section{Introduction}\label{sec:intoduction}

The use of Machine Learning (ML) has become increasingly prevalent in critical domains, from healthcare and finance to autonomous systems and safety monitoring. As ML models grow in complexity, so does the demand for their output to be interpretable by humans, a need that has given rise to the field of eXplainable Artificial Intelligence (XAI). Historical approaches to model interpretability have focused on linear models, decision trees, or post hoc interpretation techniques, but the advent of high-performing black-box models, such as deep neural networks and ensemble methods, has significantly increased the difficulty of understanding internal decision-making processes~\cite{guidotti2018survey,vassiliades2021argumentation}.

In parallel, the emergence of Large Language Models (LLMs), such as OpenAI's GPT\footnote{\url{https://openai.com/}}, Anthropic's Claude\footnote{\url{https://claude.ai/}}, and DeepSeek\footnote{\url{https://www.deepseek.com/}}, has opened up new research directions, not only in natural language understanding, but also in assisting data science workflows. Recent work suggests that LLMs can be used as AI assistants capable of writing code, suggesting ML pipelines, and automating parts of the data analysis lifecycle~\cite{koch2023teach}. However, their ability to generate solutions that are not only effective but also explainable has yet to be fully understood or systematically evaluated. Moreover, in the literature we can see that LLMs have been used successfully in various domains such as malware detection, DDOS attack detection, vulnerability detection, forecasting, among other areas \cite{guastalla2023application,nelson2025chatgpt,sheston2025finetuing,su2024large,zhou2024large}.

This paper investigates the explainability of LLM-generated ML solutions. We analyze this question through two classification tasks: a binary classification problem to predict whether a driver is in an alert or non-alert state (based on a custom-made dataset developed specifically for this research) and a multilabel classification task using the publicly available \textit{yeast} dataset\footnote{\url{https://www.kaggle.com/datasets/samanemami/yeastcsv/data}} which contains data on protein localization sites. The ML models were not hand-coded, but rather produced through prompting LLMs to generate pipelines for training four classifiers: Random Forest, XGBoost, Multi-Layer Perceptron (MLP), and Long Short-Term Memory (LSTM). We then evaluated the quality of these models in terms of both performance (recall, precision, F1-score) and explainability using SHAP-based metrics: SHAP Fidelity (mean squared error between SHAP values and true predictions) and SHAP Sparsity (average number of features with non-zero attribution).

The motivation behind this study lies in the increasing reliance on LLMs to automate ML workflows. Although these models can generate working pipelines, it remains unclear whether the results are interpretable or optimized for human understanding. Especially in domains where trust and accountability are crucial, this gap in explainability could pose serious risks.

Our key contributions are as follows:
\begin{itemize}
    \item We perform a systematic comparison of LLM-generated pipelines for binary and multilabel classification using four popular ML models.
    \item We introduce a custom driver alertness dataset, annotated for binary classification, which we make available to the research community.
    \item We evaluate not just model performance, but also model explainability using SHAP Fidelity and Sparsity, to assess the transparency of LLM-generated pipelines.
\end{itemize}

The remainder of the paper is structured as follows: Section~\ref{sec:related} reviews related work on LLMs and model explainability. Section~\ref{sec:methodology} describes the methodology for generating, training, and interpreting ML pipelines. Section~\ref{sec:evaluation} presents the results of our performance and explainability analysis. Section~\ref{sec:discussion} discusses our findings, limitations, and implications. Finally, Section~\ref{sec:conclusion} concludes the paper and outlines the directions for future research.

\section{Related Work}\label{sec:related}

XAI addresses the interpretability gap of modern black-box models through post-hoc and inherently transparent approaches \cite{guidotti2018survey,lundberg2017unified,ribeiro2016why}. Among model-agnostic tools, SHAP is widely adopted for its game-theoretic grounding and local/global utility \cite{lundberg2017unified,covert2021explaining}. In parallel, LLMs increasingly automate ML workflows \cite{liu2023survey,koch2023teach}, yet their explainability characteristics remain under-studied.

Parallel to advances in XAI, the emergence of LLMs such as OpenAI's GPT, Anthropic's Claude, and DeepSeek has transformed many domains, including code generation, automated reasoning, and scientific discovery~\cite{koch2023teach,liu2023summary}. Recent research has begun to explore the use of LLMs for automating parts of the ML lifecycle, including data preprocessing, feature engineering, model selection, and even hyperparameter tuning. Projects like AutoML-Zero~\cite{real2020automl} hint at a future where ML workflows can be automatically discovered without human intervention, and LLMs represent a powerful addition to this trend.

Despite their promise, the use of LLMs in generating interpretable ML solutions remains under explored. Early studies suggest that while LLMs can generate syntactically correct and often effective ML pipelines, their output may not always prioritize model transparency, robustness, or ethical considerations~\cite{bommasani2021opportunities}. Moreover, LLMs operate as stochastic systems---minor changes in prompts can lead to widely varying output, introducing challenges in reproducibility and explainability.

Previous work, such as Gilpin et al.~\cite{gilpin2018explaining}, emphasizes the importance of explanation faithfulness and user trust in ML applications, warning against over-relying on superficial interpretability methods. Applying such concerns to LLM-generated pipelines is crucial, given that LLMs are not inherently aware of concepts such as causality, feature attribution robustness, or bias mitigation unless explicitly guided through careful prompting.

In summary, while considerable progress has been made separately in XAI and LLM-driven ML automation, the intersection of these two fields---evaluating the explainability of ML pipelines generated by LLMs---remains relatively untouched. Our study addresses this gap by systematically evaluating the fidelity and sparsity of explanations for models generated by LLMs in two different classification settings.

\section{Methodology}\label{sec:methodology}

Here, the purpose is to describe the nature of data in Sub-section \ref{sub:nature-data}, that was used in the two use cases that we tested the LLMs. Also, to present the prompt that we used in each case, described in Sub-section \ref{sub:prompting}. We then present the pipeline of how the LLMs produced code to solve the binary and multilabel classification problems, and how those were evaluated for their level of explainability with SHAP sparsity and fidelity.

\begin{figure}[!b]
    \centering
    \includegraphics[width=0.78\linewidth]{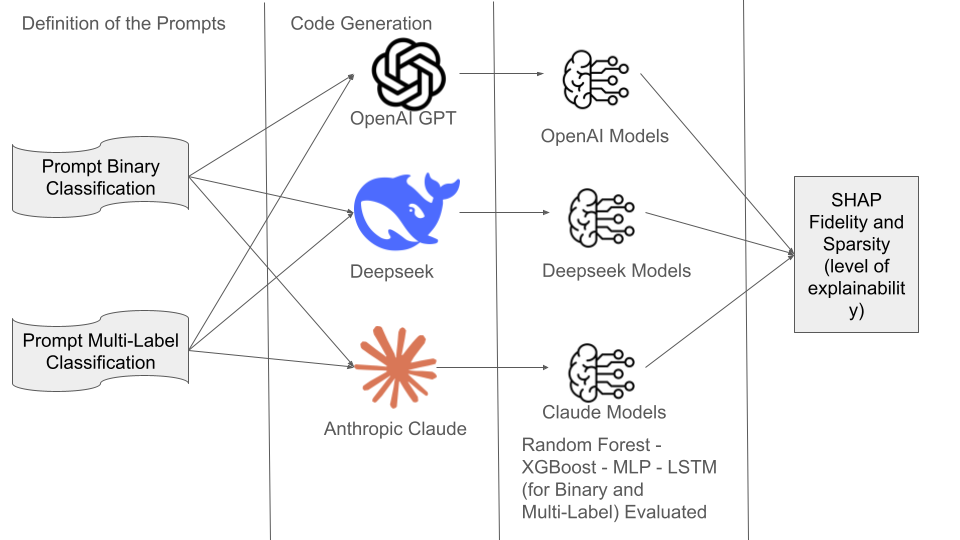}
    \caption{Overview of the experimental pipeline.}
    \label{fig:pipeline}
\end{figure}

In more detail, Figure \ref{fig:pipeline} illustrates the experimental pipeline designed to assess the explainability of LLM-generated ML workflows. The process begins with the formulation of prompts tailored to two distinct classification tasks: binary classification (for predicting driver alertness) and multilabel classification (using the yeast dataset). These prompts are then fed into three different state-of-the-art LLMs (i.e. OpenAI GPT, DeepSeek, and Anthropic Claude), which act as autonomous agents for code generation. The role of each LLM is to produce executable training pipelines that include preprocessing steps, model selection, and evaluation logic. This prompt-to-code phase reflects the emerging use of LLMs as collaborators in the ML lifecycle.

Once the code is generated, it is executed to train four common types of classifiers — Random Forest, XGBoost, MLP, and LSTM — across both classification tasks. The models produced by each LLM are then evaluated not only in terms of predictive performance (e.g., recall, precision, F1-score) but also in terms of explainability using SHAP-based metrics. SHAP Fidelity (how well the SHAP explanation reflects the model's actual outputs) and SHAP Sparsity (how concisely important features are identified) are used as quantitative indicators of interpretability. The pipeline thus enables a systematic comparison of how well different LLMs can generate ML solutions that are both effective and explainable, offering insights into the reliability of LLMs as tools for interpretable AI development.

\subsection{Nature of Data}\label{sub:nature-data}

In this study, two datasets were used to evaluate the explainability of LLM-generated ML solutions across binary and multilabel classification tasks.

The \textbf{alert dataset}, synthetically generated for drowsiness detection, contains 200k samples with five features: heart rate (continuous), yawning, straight-ahead gaze, eye closure (binary), and a binary alertness label (alert/non-alert). Balanced classes and domain-relevant features make it suitable for testing interpretability in binary classification models, a common practice in driver monitoring research~\cite{drowsy_survey}.

The \textbf{yeast dataset}, a benchmark in multilabel learning~\cite{elisseeff2001kernel}, consists of 1484 instances with eight numerical features (mcg, gvh, alm, mit, erl, pox, vac, nuc) describing protein localization in yeast cells. Samples may belong to multiple sites (e.g., mitochondria, nucleus), making it a multilabel problem. Its moderate size, multiple labels, and biological relevance make it ideal for evaluating interpretability in multilabel models.

Together, these datasets provide a framework for assessing LLMs in generating interpretable ML pipelines under classification settings of differing complexity: binary decision-making in behavioural contexts versus multilabel biological classification.
\subsection{Prompt Construction}\label{sub:prompting}

Prompt engineering plays a critical role in directing LLMs to perform specific tasks, particularly when those tasks involve generating executable code. The structure, clarity, and detail of a prompt directly influence the model's behaviour and the quality of its output. For this study, we carefully designed prompts to ensure that LLMs would understand the problem settings, data structures, and required evaluation metrics, thus generating relevant and consistent ML workflows~\cite{white2023prompt}.

We designed two prompts, one for binary and one for multilabel classification, written in natural language to describe dataset format, features, targets, and required models/metrics. The goal was to reduce ambiguity while ensuring code aligned with standard data science practice.

For the binary task, the prompt specifies a synthetic driver alertness dataset with five behavioural features and asks for training/evaluation of one of four classifiers (Random Forest, XGBoost, MLP, LSTM) using accuracy, precision, recall, and F1. For the multilabel yeast dataset, it outlines the biological context, features, and multilabel targets with the same model and evaluation requirements. Both prompts provide just enough metadata to guide LLMs without over-constraining them.

The LLMs were queried with these prompts using a temperature setting of 1.0 to encourage diverse and creative solutions, while still expecting the output to be functionally valid. Next, we include the actual prompts used in the study:

\paragraph{Prompt for Binary Classification}
\begin{quote}
\texttt{(1) chat I have a csv with 200k rows the rows are analyzed\\
(2) below heart\_rate,yawning,looks\_straight,eyes\_closed,alert\\
(3) heart\_rate = human heart rate (positive integer \\ (4) usually no more than 160)\\
(5) yawning = boolean value that indicates if the person is\\  (6) yawning in each row or not 0 means yes 1 means no\\
(7) looks\_straight = boolean value that indicates if the \\ (8) person is looking straight ahead or not \\ 
(9) , again 0 means no 1 means \\ (10) yes eyes\_closed = boolean that indicates\\
(11) if the eyes of the \\ (12) person are closed 0 means yes 1 means no\\
(13) I want you to train and evaluate (split the dataset \\ (14) randomly) an \{random forest; xgboost; mlp; lstm\} model.\\
(15) for the evaluation I want the accuracy, the precision, \\ (16) the recall and F1 score that the model achieved}
\end{quote}

\paragraph{Prompt for Multilabel Classification}
\begin{quote}
\texttt{(1) I have a CSV file called yeast.csv.\\
(2) It has 8 columns and 1484 rows, with the following\\
(3) characteristics Instances: 1,484 yeast proteins (rows)\\
(4) Features: 8 numeric features (attributes):\\
(5) mcg: McGeoch\'s method for signal sequence recognition\\
(6) gvh: von Heijne\'s method\\
(7) alm: Score for the presence of an Aliphatic region\\
(8) mit: Score for a mitochondrial targeting sequence\\
(9) erl, pox, vac, etc.\\
(10) Label (target): Protein localization site (e.g., CYT, \\ 
(11) NUC, MIT, ME1, etc.)\\
(12) Task: Multi-class classification (predict \\ (13) protein location from numeric features)\\
(14) The features are numeric, and the last  \\ (15) column is the target (protein localization site). I want \\ (16) to build a classification model using \{Random Forest; \\ (17) XGBoost; MLP; LSTM\} to predict the target. Can you help  \\ (18) me with the code to train and evaluate the model?\\
(19) for the evaluation I want the accuracy, the precision, \\ 
(20) the recall and F1 score that the model achieved}
\end{quote}

These prompts served as a stable interface between human intent and automated pipeline generation, allowing us to analyze not just the performance but also the explainability of models created under different prompt conditions and model architectures.

\subsection{Description of the Pipeline}\label{sub:pipeline}

This section details the methodology followed to assess the explainability capabilities of LLMs when tasked with generating ML pipelines. The experimental setup involved two classification tasks: a binary classification problem using a custom driver alertness dataset and a multilabel classification problem using the publicly available yeast dataset. For each task, carefully designed prompts were constructed and provided to three LLMs: OpenAI GPT, Anthropic Claude, and DeepSeek. The LLMs were asked to generate code for training and evaluating four different ML models: Random Forest, XGBoost, MLP, and LSTM networks.

Each generated script was executed to train the corresponding model, and its outputs were evaluated on two axes: predictive performance (measured by accuracy, precision, recall, and F1-score) and model explainability. Explainability was assessed using SHAP values, with two specific metrics: SHAP Fidelity (mean squared error between model outputs and SHAP approximations) and SHAP Sparsity (the number of features deemed influential). In total, for each classification task, 12 scripts were executed (4 models × 3 LLMs), resulting in a comprehensive cross-comparison of performance and explainability across generated pipelines.

To formalize our metrics, let $f(\mathbf{x})$ denote the model prediction for input $\mathbf{x}$ and $g(\mathbf{x})$ the SHAP surrogate explanation with baseline $\phi_0$ and feature attributions $\phi_i(\mathbf{x})$. Then
\[
g(\mathbf{x}) = \phi_0 + \sum_{i=1}^d \phi_i(\mathbf{x}).
\]
We define Average SHAP Fidelity as the mean squared error (MSE) between $f$ and $g$ across the dataset $\mathcal{D}$:
\[
\text{Fidelity} = \frac{1}{|\mathcal{D}|}\sum_{\mathbf{x}\in\mathcal{D}}\big(f(\mathbf{x})-g(\mathbf{x})\big)^2.
\]
Sparsity is the average number of features with non-zero attribution (above threshold $\tau$):
\[
\text{Sparsity} = \frac{1}{|\mathcal{D}|}\sum_{\mathbf{x}\in\mathcal{D}} 
\Big|\{\,i : |\phi_i(\mathbf{x})| > \tau \,\}\Big|.
\]
Low Fidelity indicates explanations that closely track model outputs, while low Sparsity indicates concise, human-readable explanations.

\subsection{Binary Classification Model Generation and Analysis}

For the binary classification task, we employed a custom driver alertness dataset of 200,000 samples with five physiological and behavioural features. LLMs were prompted to generate training and evaluation code for four classifiers (Random Forest, XGBoost, MLP, LSTM), using an 80/20 train–test split. Non-sequential models were trained in standard single-step fitting, while LSTMs followed default sequential training. No explicit optimization (e.g., cross-validation or tuning) was applied, as the objective was to assess the raw effectiveness and explainability of first-attempt pipelines produced by LLMs.

The evaluation of the model was standardized across all models using four performance metrics: accuracy, precision, recall, and F1-score. These metrics were computed on the basis of predictions made on the testing set, ensuring consistency in comparing model outputs.

After evaluation of the model, the explainability was assessed using SHAP. Specifically:
\begin{itemize}
    \item \textbf{SHAP Fidelity} was calculated as the Mean Squared Error (MSE) between the model’s predicted probabilities and the SHAP additive explanations, quantifying how well SHAP approximates the true behaviour of the model.
    \item \textbf{SHAP Sparsity} was measured as the average number of non-zero feature attributions per prediction, providing insight into the simplicity and interpretability of the model's decision-making process.
\end{itemize}

A comprehensive summary of the LLM-generated pipelines for the binary classification task is presented in Table~\ref{tab:binary_summary}.

\begin{table}[!ht]
    \centering
    \caption{Summary of binary classification model generation across LLMs. \\
    Accuracy (A), Precision (P), Recall (R), random split (rs) and epoch (e)}
    \label{tab:binary_summary}
    \begin{tabular}{|l|l|l|l|}
    \hline
    \textbf{LLM} & \textbf{Model} & \textbf{Metrics Computed} & \textbf{Training Details} \\
    \hline
    OpenAI    & Random Forest  & A, P, R, F1 & 80/20 rs \\
    OpenAI    & XGBoost         & A, P, R, F1  & 80/20 rs \\
    OpenAI    & MLP             & A, P, R, F1  & 80/20 rs \\
    OpenAI    & LSTM            & A, P, R, F1  & 80/20 rs, 20 e \\
    \hline
    DeepSeek  & Random Forest   & A, P, R, F1  & 80/20 rs \\
    DeepSeek  & XGBoost         & A, P, R, F1  & 80/20 rs \\
    DeepSeek  & MLP             & A, P, R, F1  & 80/20 rs \\
    DeepSeek  & LSTM            & A, P, R, F1  & 80/20 rs, 20 e \\
    \hline
    Claude    & Random Forest   & A, P, R, F1  & 80/20 rs \\
    Claude    & XGBoost         & A, P, R, F1  & 80/20 rs \\
    Claude    & MLP             & A, P, R, F1  & 80/20 rs \\
    Claude    & LSTM            & A, P, R, F1  & 80/20 rs, 20 e \\
    \hline
    \end{tabular}
\end{table}

\subsection{Multilabel Classification Model Analysis}

For the multilabel classification task, the yeast dataset was used, containing 1,484 samples with eight numerical features and multilabel targets for protein localization sites. As in the binary setup, LLMs generated pipelines for Random Forest, XGBoost, MLP, and LSTM models. Tree-based models (Random Forest, XGBoost) used \texttt{OneVsRestClassifier} wrappers to decompose the problem into binary tasks, while neural models (MLP, LSTM) incorporated multiple output neurons with sigmoid activations and binary cross-entropy loss. Training employed an 80/20 random split, and evaluation included accuracy, precision, recall, and F1-score. SHAP-based fidelity and sparsity analyses were applied in line with the binary classification procedure.

A comprehensive summary of the LLM-generated pipelines for the multilabel classification task is presented in Table~\ref{tab:multilabel_summary}.

\begin{table}[!th]
    \centering
    \caption{Summary of multilabel classification model generation across LLMs. \\ Accuracy (A), Precision (P), Recall (R), random split (rs) and epoch (e)}
    \label{tab:multilabel_summary}
    \begin{tabular}{|l|l|l|l|}
    \hline
    \textbf{LLM} & \textbf{Model} & \textbf{Metrics Computed} & \textbf{Training Details} \\
    \hline
    OpenAI    & Random Forest  & A, P, R, F1 & 80/20 rs, OneVsRest \\
    OpenAI    & XGBoost         & A, P, R, F1 & 80/20 rs, OneVsRest \\
    OpenAI    & MLP             & A, P, R, F1 & 80/20 rs, multilabel output \\
    OpenAI    & LSTM            & A, P, R, F1 & 80/20 rs, multilabel output, 20 e \\
    \hline
    DeepSeek  & Random Forest   & A, P, R, F1 & 80/20 rs, OneVsRest \\
    DeepSeek  & XGBoost         & A, P, R, F1 & 80/20 rs, OneVsRest \\
    DeepSeek  & MLP             & A, P, R, F1 & 80/20 rs, multilabel output \\
    DeepSeek  & LSTM            & A, P, R, F1 & 80/20 rs, multilabel output, 20 e \\
    \hline
    Claude    & Random Forest   & A, P, R, F1 & 80/20 rs, OneVsRest \\
    Claude    & XGBoost         & A, P, R, F1 & 80/20 rs, OneVsRest \\
    Claude    & MLP             & A, P, R, F1 & 80/20 rs, multilabel output \\
    Claude    & LSTM            & A, P, R, F1 & 80/20 rs, multilabel output, 20 e \\
    \hline
    \end{tabular}
\end{table}

\section{Evaluation}\label{sec:evaluation}

Model evaluation in ML typically involves assessing both predictive performance and model interpretability. For predictive performance, we computed three widely accepted metrics: \textbf{precision}, \textbf{recall}, and \textbf{F1-score}. Precision reflects the proportion of correctly predicted positive observations to the total predicted positives, making it critical in domains where false positives are costly. Recall measures the proportion of correctly predicted positives out of all actual positives, crucial for high-sensitivity applications. The F1-score represents the harmonic mean of precision and recall, offering a balanced measure when both false positives and false negatives are significant~\cite{powers2020evaluation}.

Beyond performance, we evaluated the \textbf{explainability} of each model using SHAP~\cite{lundberg2017unified}. Two key explainability metrics were employed: \textbf{SHAP Fidelity} and \textbf{SHAP Sparsity}. SHAP Fidelity, measured as the MSE between the model outputs and SHAP approximations, indicates how faithfully the SHAP explanations represent the true model behaviour. Lower values imply that the explanations are highly accurate reflections of the model's decision process. SHAP Sparsity measures the average number of features with non-zero importance per prediction; a lower sparsity suggests simpler, more focused explanations that enhance human interpretability. High sparsity (many features) may indicate over complex explanations, while low sparsity points to streamlined, understandable decision mechanisms.

It is important to note that the \textbf{custom alertness dataset} used in the binary classification task exhibited \textbf{largely linear relationships} between the input features and the target label, as verified by correlation analysis. This inherent simplicity facilitated the models to achieve very high predictive performance (precision, recall, and F1 all close to 1.0), and produce clean SHAP explanations with perfect or near-perfect fidelity and consistent sparsity levels. Consequently, the observed results reflect both the effectiveness of LLM-generated pipelines and the learnability of the underlying task structure.

The numerical results for performance and explainability metrics for all models and LLMs are summarized below. In Tables \ref{tab:performance_summary} and \ref{tab:explainability_summary}, we also provide the baseline score (i.e., on code that was written by the authors and trained on the same principles).

\begin{table}[!t]
    \centering
    \caption{Performance metrics: Precision, Recall, and F1-Score across LLMs and Models}
    \label{tab:performance_summary}
    \begin{tabular}{|l|l|c|c|c|}
    \hline
    \textbf{LLM} & \textbf{Model} & \textbf{~Precision~} & \textbf{~Recall~} & \textbf{~F1-Score~} \\
    \hline
    OpenAI   & Random Forest & 1.0000 & 1.0000 & 1.0000 \\
    OpenAI   & XGBoost        & 1.0000 & 1.0000 & 1.0000 \\
    OpenAI   & MLP            & 1.0000 & 1.0000 & 1.0000 \\
    OpenAI   & LSTM           & 0.9984 & 1.0000 & 0.9992 \\
    \hline
    DeepSeek & Random Forest  & 1.0000 & 1.0000 & 1.0000 \\
    DeepSeek & XGBoost        & 1.0000 & 1.0000 & 1.0000 \\
    DeepSeek & MLP            & 1.0000 & 1.0000 & 1.0000 \\
    DeepSeek & LSTM           & 0.9967 & 1.0000 & 0.9970 \\
    \hline
    Claude   & Random Forest  & 1.0000 & 1.0000 & 1.0000 \\
    Claude   & XGBoost        & 1.0000 & 1.0000 & 1.0000 \\
    Claude   & MLP            & 0.9984 & 1.0000 & 0.9992 \\
    Claude   & LSTM           & 1.0000 & 1.0000 & 1.0000 \\
    \hline
    Baseline   & Random Forest  & 1.0000 & 1.0000 & 1.0000 \\
    Baseline   & XGBoost        & 1.0000 & 1.0000 & 1.0000 \\
    Baseline   & MLP            & 1.0000 & 1.0000 & 1.0000 \\
    Baseline   & LSTM           & 1.0000 & 1.0000 & 1.0000 \\
    \hline
    \end{tabular}
\end{table}

\begin{table}[!b]
    \centering
    \caption{Explainability metrics: SHAP Fidelity and Sparsity across LLMs and Models}
    \label{tab:explainability_summary}
    \resizebox{1\textwidth}{!}{
    \begin{tabular}{|l|l|c|c|}
    \hline
    \textbf{LLM} & \textbf{Model} & \textbf{SHAP Fidelity MSE} & \textbf{SHAP Sparsity (Avg Features)} \\
    \hline
    OpenAI   & Random Forest  & 0.00000 & 4.00 \\
    OpenAI   & XGBoost        & 0.00000 & 4.00 \\
    OpenAI   & MLP            & 0.00000 & 4.00 \\
    OpenAI   & LSTM           & 0.01000 & 4.00 \\
    \hline
    DeepSeek & Random Forest  & 0.00000 & 4.00 \\
    DeepSeek & XGBoost        & 0.00000 & 4.00 \\
    DeepSeek & MLP            & 0.00000 & 4.00 \\
    DeepSeek & LSTM           & 0.05994 & 4.00 \\
    \hline
    Claude   & Random Forest  & 0.00000 & 4.00 \\
    Claude   & XGBoost        & 0.00000 & 4.00 \\
    Claude   & MLP            & 0.02810 & 4.00 \\
    Claude   & LSTM           & 0.00000 & 4.00 \\
    \hline
    Baseline   & Random Forest  & 0.00000 & 4.00 \\
    Baseline   & XGBoost        & 0.00000 & 4.00 \\
    Baseline   & MLP            & 0.00000 & 4.00 \\
    Baseline   & LSTM           & 0.00000 & 4.00 \\
    \hline
    \end{tabular}
    }
\end{table}

For the multilabel classification task based on the \textit{yeast} dataset, we evaluated the generated models using the same metrics: precision, recall, and F1-score. However, compared to the binary classification setting, the multilabel task inherently presents more complexity. Each instance may belong to one of ten possible classes, and class imbalance further complicates the learning process. Therefore, lower performance metrics are expected and natural.

In addition to predictive performance, the model explainability was again assessed through SHAP Fidelity and Sparsity. Despite the greater classification difficulty, SHAP Fidelity values remained extremely high (close to 0), suggesting that the generated explanations continue to accurately reflect model behaviour. Sparsity values ranged from approximately 6.9 to 8.0, indicating that for each prediction, a compact subset of features was influential, maintaining a desirable level of interpretability even in a more challenging setting.

Training times remained reasonable due to the moderate size of the yeast dataset (1,484 samples, 8 features), and the complexity of the model was manageable without specialized hardware acceleration. The detailed performance and explainability results for all LLM-generated pipelines are summarized below. In Tables \ref{tab:multilabel_performance_summary} and \ref{tab:multilabel_explainability_summary}, we also provide the baseline score (i.e., on code that was written by the authors and trained on the same principles).

\begin{table}[!b]
    \centering
    \caption{Performance metrics: Multilabel classification}
    \label{tab:multilabel_performance_summary}
    \begin{tabular}{|l|l|c|c|c|c|}
    \hline
    \textbf{LLM} & \textbf{Model} & \textbf{~Accuracy~} & \textbf{~Precision~} & \textbf{~Recall~} & \textbf{~F1-Score~} \\
    \hline
    OpenAI   & Random Forest  & 0.6162 & 0.5329 & 0.4469 & 0.4636 \\
    OpenAI   & XGBoost        & 0.5791 & 0.5860 & 0.5041 & 0.5203 \\
    OpenAI   & MLP            & 0.5488 & 0.6215 & 0.5465 & 0.5582 \\
    OpenAI   & LSTM           & 0.5859 & 0.5991 & 0.5311 & 0.5417 \\
    \hline
    DeepSeek & Random Forest  & 0.6263 & 0.6143 & 0.6263 & 0.6178 \\
    DeepSeek & XGBoost        & 0.6061 & 0.5969 & 0.6061 & 0.5993 \\
    DeepSeek & MLP            & 0.6364 & 0.6333 & 0.6364 & 0.6322 \\
    DeepSeek & LSTM           & 0.6128 & 0.6049 & 0.6128 & 0.6071 \\
    \hline
    Claude   & Random Forest  & 0.6330 & 0.6305 & 0.6330 & 0.6242 \\
    Claude   & XGBoost        & 0.6027 & 0.5994 & 0.6027 & 0.5976 \\
    Claude   & MLP            & 0.6364 & 0.6333 & 0.6364 & 0.6322 \\
    Claude   & LSTM           & 0.6128 & 0.6049 & 0.6128 & 0.6071 \\
    \hline
    Baseline   & Random Forest  & 0.6330 & 0.6305 & 0.6330 & 0.6242 \\
    Baseline   & XGBoost        & 0.6061 & 0.5969 & 0.6061 & 0.5993  \\
    Baseline   & MLP            & 0.6162 & 0.5329 & 0.4469 & 0.4636 \\
    Baseline   & LSTM           & 0.6364 & 0.6333 & 0.6364 & 0.6322  \\
    \hline
    \end{tabular}
\end{table}

\begin{table}[!h]
    \centering
    \caption{Explainability metrics: Multilabel classification}
    \label{tab:multilabel_explainability_summary}
    \resizebox{\textwidth}{!}{
    \begin{tabular}{|l|l|c|c|}
    \hline
    \textbf{LLM} & \textbf{Model} & \textbf{SHAP Fidelity MSE} & \textbf{SHAP Sparsity (Avg Features)} \\
    \hline
    OpenAI   & Random Forest & 0.00000 & 7.98 \\
    OpenAI   & XGBoost        & 0.00000 & 8.00 \\
    OpenAI   & MLP            & 0.00000 & 8.00 \\
    OpenAI   & LSTM           & 0.00000 & 8.00 \\
    \hline
    DeepSeek & Random Forest  & 0.00000 & 7.00 \\
    DeepSeek & XGBoost        & 0.00000 & 7.01 \\
    DeepSeek & MLP            & 0.00000 & 6.90 \\
    DeepSeek & LSTM           & 0.00000 & 6.90 \\
    \hline
    Claude   & Random Forest  & 0.00000 & 8.00 \\
    Claude   & XGBoost        & 0.00000 & 8.00 \\
    Claude   & MLP            & 0.00000 & 6.90 \\
    Claude   & LSTM           & 0.00000 & 6.90 \\
    \hline
    Baseline   & Random Forest  & 0.00000 & 8.00 \\
    Baseline   & XGBoost        & 0.00000 & 8.00 \\
    Baseline   & MLP            & 0.00000 & 7.00 \\
    Baseline   & LSTM           & 0.00000 & 6.90 \\
    \hline
    \end{tabular}
    }
\end{table}

Overall, the results highlight that even under multilabel settings, LLM-generated ML pipelines can achieve not only reasonable predictive performance but also extremely faithful and sparse explanations. The slight reduction in sparsity compared to the binary classification case reflects the increased complexity and number of output classes, but remains within acceptable levels to ensure interpretability.

All models were trained and evaluated on Google Colab, using a virtualized environment composed of an Intel Xeon CPU, approximately 12 GB RAM, and no dedicated GPU for the experiments.

\section{Discussion}\label{sec:discussion}

This study systematically evaluated the explainability of LLM-generated ML pipelines across two tasks: binary driver alertness classification and multilabel protein localization, using three state-of-the-art LLMs (OpenAI GPT, DeepSeek, Anthropic Claude).

Results show that LLMs can autonomously produce effective ML pipelines with minimal human input. In the binary task, all models achieved near-perfect precision, recall, and F1 scores, reflecting the dataset’s linear structure. SHAP analysis confirmed high fidelity (almost zero error) and low sparsity (~4 features per prediction), ensuring concise and accurate explanations. In the more complex multilabel task, performance dropped (F1 = 0.5-0.63), yet pipelines remained competitive, with DeepSeek and Claude slightly outperforming GPT. SHAP fidelity stayed perfect, with sparsity at acceptable levels (6.9-8.0 features).

We also observe systematic differences among LLMs and model families. Claude and DeepSeek tended to produce pipelines with slightly more consistent multilabel handling (e.g., explicit probability stacking), which yielded marginally higher macro-F1 than GPT.  By contrast, GPT often defaulted to shallower architectures or fewer estimators unless explicitly prompted.  Across models, tree ensembles (RF, XGBoost) consistently produced sparser explanations, while MLPs showed slightly higher variance in sparsity. LSTMs, although generated by all LLMs, did not provide additional benefit in these tabular settings, highlighting the importance of model choice beyond code generation.

Limitations include testing only three LLMs, a narrow set of ML models, reliance on SHAP as the sole explainability metric, and single-shot prompting without optimization. Broader datasets, alternative interpretability methods, and more systematic prompting are needed for generalization.

Overall, the findings demonstrate that LLMs can generate interpretable, high-performing ML workflows. Future research should enhance robustness, fairness, and causal transparency to strengthen their use in automated ML pipeline generation.
\section{Conclusion}\label{sec:conclusion}

This work presents an initial investigation into the ability of LLMs to autonomously generate ML pipelines that are not only effective, but also explainable. By formulating clear prompts for binary and multilabel classification tasks, and evaluating the resulting models across both predictive performance and SHAP-based explainability metrics, we demonstrate that LLMs like OpenAI GPT, Anthropic Claude, and DeepSeek can reliably produce interpretable ML solutions. Our results show that these models can approximate standard workflows in data science and yield outputs that are faithful and sparse in terms of feature attribution—qualities aligned with the goals of explainable AI.

However, this study also underscores current limitations in prompt-based ML automation, including inconsistencies between LLM outputs and the narrow scope of interpretability assessment through SHAP alone. Future work could extend prompting to complex ML tasks like regression and time series forecasting, benchmark LLM pipelines against human baselines, and explore causal explainability and fairness-aware evaluations to deepen trust in LLM-driven workflows.

\section*{Acknowledgments}
Funded by the European Union Horizon Europe programme through ALFIE under Grant Agreement 101177912.


\begin{thebibliography}{99}

\bibitem{bommasani2021opportunities}
Bommasani, R., Hudson, D.A., Adeli, E., et al. (2021). On the opportunities and risks of foundation models. {arXiv:2108.07258}.

\bibitem{covert2021explaining}
Covert, I., Lundberg, S.M., \& Lee, S.-I. (2021). Explaining by removing: A unified framework for model explanation. \textit{Journal of Machine Learning Research}, 22(209), 1--90.

\bibitem{elisseeff2001kernel}
Elisseeff, A., \& Weston, J. (2001). A kernel method for multi-labelled classification. In \textit{Advances in Neural Information Processing Systems 14} (pp. 681--687).

\bibitem{gilpin2018explaining}
Gilpin, L.H., Bau, D., Yuan, B.Z., et al.
(2018). Explaining explanations: An overview of interpretability of machine learning. In \textit{Proceedings of the 5th IEEE International Conference on Data Science and Advanced Analytics (DSAA)} (pp. 80--89).

\bibitem{guastalla2023application}
Guastalla, M., Li, Y., Hekmati, A., et al.
(2023). Application of large language models to DDOS attack detection. In \textit{Proceedings of the 1st EAI International Conference on Security and Privacy in Cyber-Physical Systems and Smart Vehicles (SmartSP)} (pp. 83-99).

\bibitem{guidotti2018survey}
Guidotti, R., Monreale, A., Ruggieri, S., et al.
(2018). A survey of methods for explaining black box models. \textit{ACM Computing Surveys}, 51(5), 1-42.

\bibitem{koch2023teach}
Koch, M., Kehlbeck, R., Friedrich, F., et al. (2023). Teach me how to prompt: A taxonomy of prompt patterns in large language models. {arXiv:2304.09810}.

\bibitem{liu2023summary}
Liu, P., Yuan, W., Fu, J., et al.
(2023). A survey of large language models. {arXiv:2303.18223}.

\bibitem{lundberg2017unified}
Lundberg, S.M., \& Lee, S.-I. (2017). A unified approach to interpreting model predictions. In \textit{Advances in Neural Information Processing Systems 30} (pp. 4765--4774).

\bibitem{drowsy_survey}
May, J.F., \& Baldwin, C L. (2009). Driver fatigue: The importance of identifying causal factors of fatigue when considering detection and countermeasure technologies. \textit{Transportation Research Part F: Traffic Psychology and Behaviour}, 12(3), 218-224.

\bibitem{nelson2025chatgpt}
Nelson, J., Pavlidis, M., Fish, A., et al.
(2025). ChatGPT-driven machine learning code generation for Android malware detection. \textit{The Computer Journal}, 68(4), 331-345.

\bibitem{powers2020evaluation}
Powers, D.M.W. (2020). Evaluation: From precision, recall and F-measure to ROC, informedness, markedness and correlation. {arXiv:2010.16061}.

\bibitem{real2020automl}
Real, E., Aggarwal, A., Huang, Y., et al.
(2020). AutoML-Zero: Evolving machine learning algorithms from scratch. In \textit{Proceedings of the 37th International Conference on Machine Learning (ICML)} (pp. 8007--8019).

\bibitem{ribeiro2016should}
Ribeiro, M.T., Singh, S., \& Guestrin, C. (2016). "Why should I trust you?": Explaining the predictions of any classifier. In \textit{Proceedings of the 22nd ACM SIGKDD International Conference on Knowledge Discovery and Data Mining (KDD)} (pp. 1135--1144).

\bibitem{sheston2025finetuing}
Shestov, A., Levichev, R., Mussabayev, R., et al.
(2025). Finetuning large language models for vulnerability detection, \textit{IEEE Access}, 13, 38889--38900.

\bibitem{su2024large}
Su, J., Jiang, C., Jin, X., et al.
(2024). Large language models for forecasting and anomaly detection: A systematic literature review. {arXiv:2402.10350}.

\bibitem{vassiliades2021argumentation}
Vassiliades, A., Bassiliades, N., \& Patkos, T. (2021). Argumentation and explainable artificial intelligence: A survey. \textit{The Knowledge Engineering Review}, 36, e5.

\bibitem{white2023prompt} 
White, T., Lin, S., Lin, Z., et al. (2023). Prompt engineering techniques for large language models: A survey. {arXiv:2302.11382}.

\bibitem{zhou2024large}
Zhou, X., Zhang, T., \& Lo, D. (2024). Large language model for vulnerability detection: Emerging results and future directions. In \textit{Proceedings of the 44th ACM/IEEE International Conference on Software Engineering: New Ideas and Emerging Results (NIER@ICSE)} (pp. 47-51).

\end{thebibliography}
\end{document}